\documentclass{article}
\usepackage{placeins}
\usepackage{caption}
\usepackage{xspace}
\usepackage{amsmath}
\usepackage{amssymb}
\usepackage{graphicx}
\usepackage{float}
\usepackage{booktabs}
\usepackage{subcaption}
\usepackage{wrapfig}

\usepackage{algorithm}
\usepackage{algorithmic}
\usepackage{enumitem}

\usepackage[preprint]{corl_2025} 
\definecolor{pmcolor}{RGB}{70,70,70}

\newcommand{\xxnote}[3]{}
\ifx\hidenotes\undefined
  \renewcommand{\xxnote}[3]{\color{#2}{#1: #3}}
\fi

\title{Feel the Force: \\Contact-Driven Learning from Humans}

\author{
  Ademi Adeniji$^{12}$\thanks{Correspondence to Ademi Adeniji: \texttt{ademiadeniji7@gmail.com}}
  \qquad
  Zhuoran Chen$^{3*}$
  \qquad
  Vincent Liu$^1$
  \AND
  Venkatesh Pattabiraman$^1$
  \qquad
  Raunaq Bhirangi$^1$
  \qquad
  Siddhant Haldar$^1$
  \AND
  Pieter Abbeel$^2$
  \qquad
  Lerrel Pinto$^1$
  \AND
  {\normalfont $^1$New York University}
  \qquad
  {\normalfont $^2$UC Berkeley}
  \qquad
  {\normalfont $^3$New York University Shanghai} \\~\\
  {\normalfont $^*$Equal contribution}
}

\newcommand{\method}{\textsc{FTF}\xspace}
\newcommand{\website}{\url{https://feel-the-force-ftf.github.io}}

\begin{document}

\maketitle


\begin{abstract}
    Controlling fine-grained forces during manipulation remains a core challenge in robotics. While robot policies learned from robot-collected data or simulation show promise, they struggle to generalize across the diverse range of real-world interactions. Learning directly from humans offers a scalable solution, enabling demonstrators to perform skills in their natural embodiment and in everyday environments. However, visual demonstrations alone lack the information needed to infer precise contact forces. We present \textsc{FeelTheForce} (\method{}): a robot learning system that models human tactile behavior to learn force-sensitive manipulation. Using a tactile glove to measure contact forces and a vision-based model to estimate hand pose, we train a closed-loop policy that continuously predicts the forces needed for manipulation. This policy is re-targeted to a Franka Panda robot with tactile gripper sensors using shared visual and action representations. At execution, a PD controller modulates gripper closure to track predicted forces—enabling precise, force-aware control. Our approach grounds robust low-level force control in scalable human supervision, achieving a 77\% success rate across 5 force-sensitive manipulation tasks. Code and videos are available at \website{}.
\end{abstract}

\keywords{Learning from Touch, Learning from Humans, Imitation Learning} 



\begin{figure}[H]
    \centering
    \includegraphics[width=\linewidth]{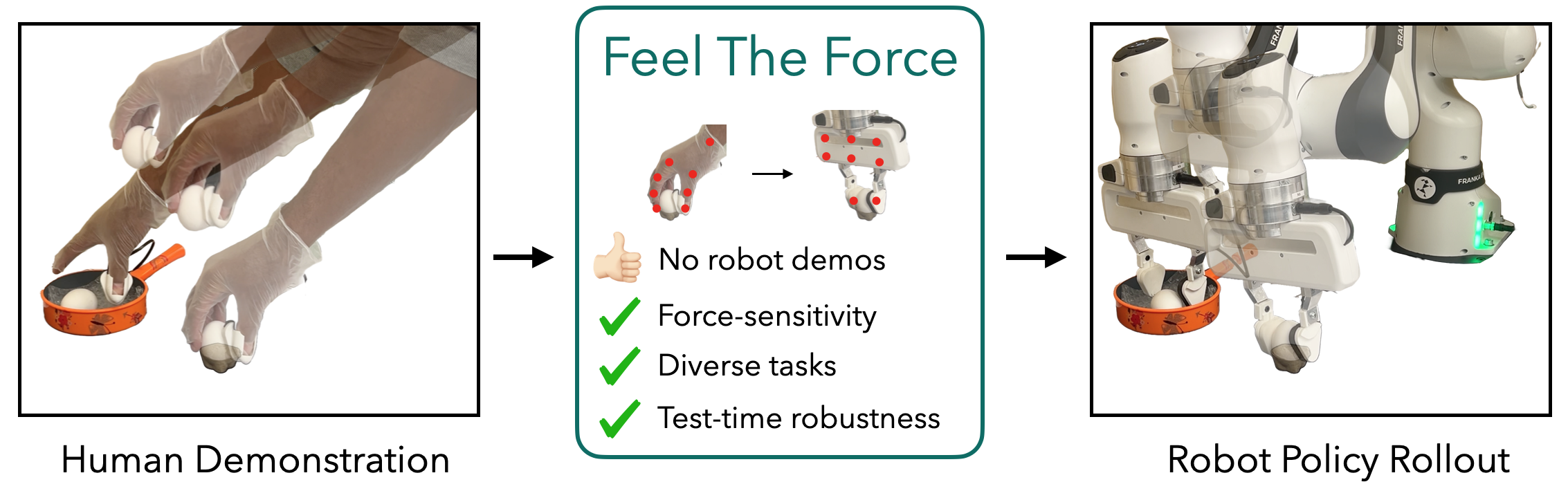} 
    \caption{$\method{}$ allows zero-shot transfer of tactile human demonstrations to a Franka Robot.}
    \label{fig:ftf_teaser}
\end{figure}

\section{Introduction}

Humans excel at manipulating the physical world not only through vision but also through a rich and nuanced sense of touch. Everyday actions—like delicately placing an egg in a bowl or unstacking a cup—depend on the ability to modulate fine-grained contact forces in real time. This form of low-level contact reasoning remains a core unsolved challenge in robotics \cite{xie2025forcefulroboticfoundationmodels}. Although recent advances in vision-based and proprioceptive imitation learning have enabled robots to perform increasingly complex tasks, they often fail when subtle force adjustments are needed, particularly under partial observability or contact uncertainty \cite{sferrazza2023powersensesgeneralizablemanipulation}. This gap arises from the mismatch between the high-bandwidth, tactile-rich control humans employ and the sparse, delayed signals available to most robotic systems.

Prior work in tactile imitation learning \cite{pattabiraman2024learningprecisecontactrichmanipulation} has explored ways to integrate force feedback into robot control. However, such work relies heavily on teleoperation which is hard to scale to diverse real-world environments \cite{khazatsky2024droidlargescaleinthewildrobot}. Furthermore, receiving tactile feedback through data-collection interfaces requires expensive haptic feedback setups \cite{liu2025vitaminlearningcontactrichtasks} or the demonstrator adjusting continuous gripper closure based on visual cues such as object deformation \cite{zhao2023learning}. In contrast, humans interact with the physical world constantly, providing a vast, underutilized source of contact-rich demonstrations in everyday settings. Works such as \cite{wang2024dexcapscalableportablemocap} propose mobile data collection systems involving glove-based motion capture. However, without force-sensing, such methods rely on task-specific priors, such as manually offsetting desired actions to approximate increased target forces. 

Even if we can obtain tactile information, a number of learning challenges stand in the way of leveraging this data for robust force-sensitive manipulation. Many methods propose to simply feed observed tactile information into the policy \cite{yu2025mimictouchleveragingmultimodalhuman, liu2025vitaminlearningcontactrichtasks, sferrazza2023powersensesgeneralizablemanipulation, pattabiraman2024learningprecisecontactrichmanipulation}. However, this passive use of tactile information suffers from two main issues. Firstly, the learned model is highly reliant on the observed force distribution. Due to the embodiment gap between the human and the robot, the executable force distributions differ. If during deployment, the robot leaves the force distribution determined by the human training data, the policy will fail to generalize. Secondly, since more information is being fed into the policy in order to predict more precise actions, much more data is often required reducing learning efficiency.

In this work, we ask: Can we endow robots with robust, force-aware control by learning efficiently from human tactile experiences? We present \textsc{FeelTheForce} (\method): a novel framework that bridges human tactile demonstrations and robotic force-sensitive manipulation. \method models human tactile-proprioceptive signals during manipulation and trains a closed-loop imitation learning policy to predict hand trajectories and desired contact forces. We collect data using a tactile glove and train a transformer-based policy that outputs hand trajectories, which are retargeted to robot end-effector poses, along with the contact force applied by the human at each timestep. At deployment, a low-level PD controller modulates the gripper closure to track the predicted force, enabling precise force reproduction across tasks without any robot data during training. This inference-time, PD-based controller continuously adjusts the robot’s behavior to stabilize around desired tactile feedback—enabling robust execution even under morphological mismatch or sensor noise. Unlike prior approaches that require action-aligned teleoperation, our method decouples the learning and execution phases, leveraging human expertise to guide robot force control.

This formulation offers two key advantages: (1) it eliminates the need for large-scale robot data and expensive haptic teleoperation and (2) it enables generalization from the human embodiment to the robot embodiment to solve force-sensitive tasks robustly. We transfer the learned policy to a Franka Panda robot with fingertip tactile sensors and evaluate on 5 force-sensitive manipulation tasks.

In summary, we demonstrate that:

\begin{itemize}[leftmargin=2em] 
\item \method robustly solves all 5 force-sensitive tasks evaluated with a 77\% success rate where baselines fail showing that active force prediction and reproduction is more effective than passive use of multi-modal force inputs.
\item \method achieves higher success rates than baselines trained on robot teleoperation data showing that the natural data collection enabled by the tactile glove can be effective for tactile data collection.
\item \method is able to achieve a success rate of 67\% on a task with adversarial disturbances during deployment, displaying robustness to test-time shifts in the tactile data distribution.
\end{itemize}

\section{Related Work}
\textbf{Tactile Sensing and Haptic Learning.} Tactile sensing plays a crucial role in enabling dexterous manipulation, particularly in tasks requiring fine-grained force control. Prior work has explored learning policies directly from tactile inputs, using sensors such as GelSight \cite{yuangelsight}, BioTac \cite{biotac}, and other vision-based tactile arrays \cite{digit}. These methods have enabled tasks like slip detection \cite{calandra2025feelingsuccessdoestouch}, grasp stability prediction \cite{stabilize}, and in-hand manipulation \cite{pan2023inhandmanipulationunknownobjects}. However, most tactile learning approaches rely on robot-collected data, which can be costly to obtain and may not generalize well across tasks or hardware. In contrast, our method leverages human tactile signals alone for training and sidesteps the need for robot interaction data.

\textbf{Imitation Learning from Human Demonstrations.} Learning from human demonstrations has a long history in robotics, encompassing techniques like behavior cloning \cite{alvinn}, inverse reinforcement learning \cite{irl}, and more recent deep imitation frameworks \cite{deepimitation, dasari2021transformers}. Most approaches rely on visual or kinematic supervision, requiring aligned action spaces or kinesthetic teaching. Some methods attempt to use wearable devices like motion capture suits or teleoperation rigs \cite{wang2024dexcapscalableportablemocap, zhao2023learning}, but these are often expensive or intrusive. While a few studies explore imitation from human force data \cite{jain2013improving, wiste2011design}, they typically assume action-space correspondence or require ground-truth contact forces. Our work is distinguished by using raw tactile and proprioceptive human signals, without requiring direct imitation of motor actions or precise temporal alignment between human and robot data.

\textbf{Human to Robot Embodiment Transfer} Transferring skills across embodiments—especially from human to robot—is a core challenge in imitation learning. Recent work \cite{guzey2024bridginghumanrobotdexterity} introduces fingertip kinematic retargeting to map human video demonstrations to coarse robot actions, which are subsequently refined through reinforcement learning. Other approaches \cite{bahl2022human, bahl2022humantorobotimitationwild} focus on extracting manipulation affordances from human videos to define high-level interaction goals for robot control. \cite{yu2025mimictouchleveragingmultimodalhuman, chi2024universal, liu2025vitaminlearningcontactrichtasks, shafiullah2023bringing} leverage hand-held data collection tools with grippers designed to match the target robot's embodiment, improving correspondence between human demonstrations and robot execution. Our method builds upon \cite{haldar2024bakuefficienttransformermultitask} by using visual hand keypoints to bridge the morphological gap between humans and robots, enabling data collection in the demonstrator’s natural embodiment while still recovering precise, executable actions for the robot.

\begin{figure}[t]
    \centering
    \includegraphics[width=\linewidth]{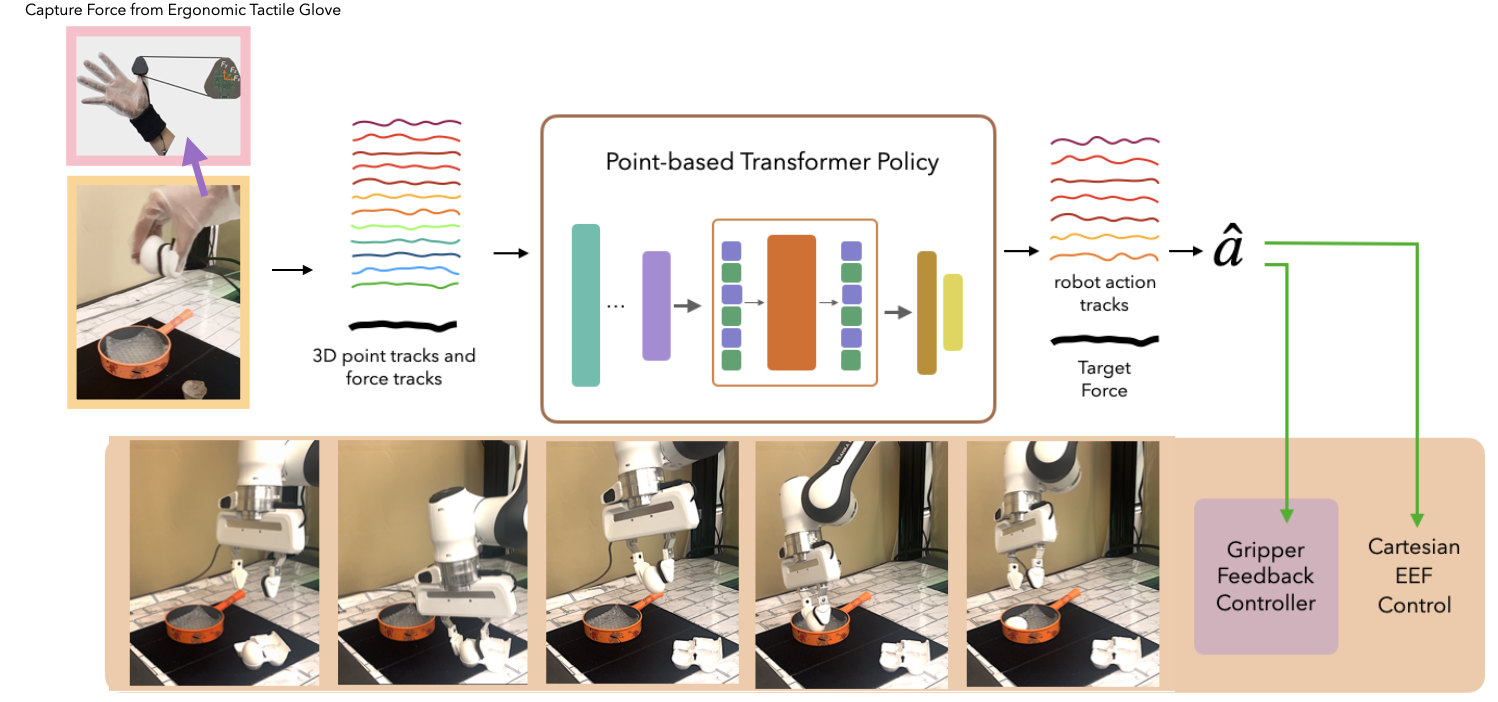} 
    \caption{$\method{}$ allows zero-shot transfer of tactile human demonstrations to a Franka Robot.}
    \label{fig:ftf_method}
\end{figure}

\section{\textsc{FeelTheForce}}
Learning force-sensitive manipulation from humans is challenging due to the complexity of inferring fine-grained contacts and forces solely from visual observations. To address this, we introduce \method{}, a framework for collecting tactile data from human demonstrations using a low-cost force-sensing glove and learning policies that predict both actions and desired forces from combined visual and tactile inputs. \method{} enables the human-to-robot embodiment transfer through a key point based unified observation and action space~\cite{haldar2025pointpolicyunifyingobservations}, while allowing human-like force modulation at inference through PD control on predicted forces.

\paragraph{Assumptions:} $(1)$ The pose of the human hand in the first frame is the same as that of the robot gripper at reset. This can be relaxed by initializing the robot to arbitrary human hand poses, which we do not investigate in this work. $(2)$ We operate in a calibrated scene where the intrinsic and extrinsic matrices for each camera is known. In practice, this is a one-time process that only takes a few minutes when the robot system is set up.



\subsection{Data Acquisition for Human-to-Robot Force Transfer}
To collect human demonstrations, \method{} enables task execution through natural human movements. During data acquisition, as the human performs the task, two calibrated RealSense cameras record visual observations of the hand and environment. At deployment, the same camera setup monitors a Franka Panda arm in the same environment. A custom tactile glove is designed to collect force data from human demonstrations, which is transferred to the robot using gripper-mounted tactile sensors. 

\begin{figure}[t]
    \centering
    \begin{subfigure}[t]{0.45\linewidth}
        \centering
        \includegraphics[width=\linewidth]{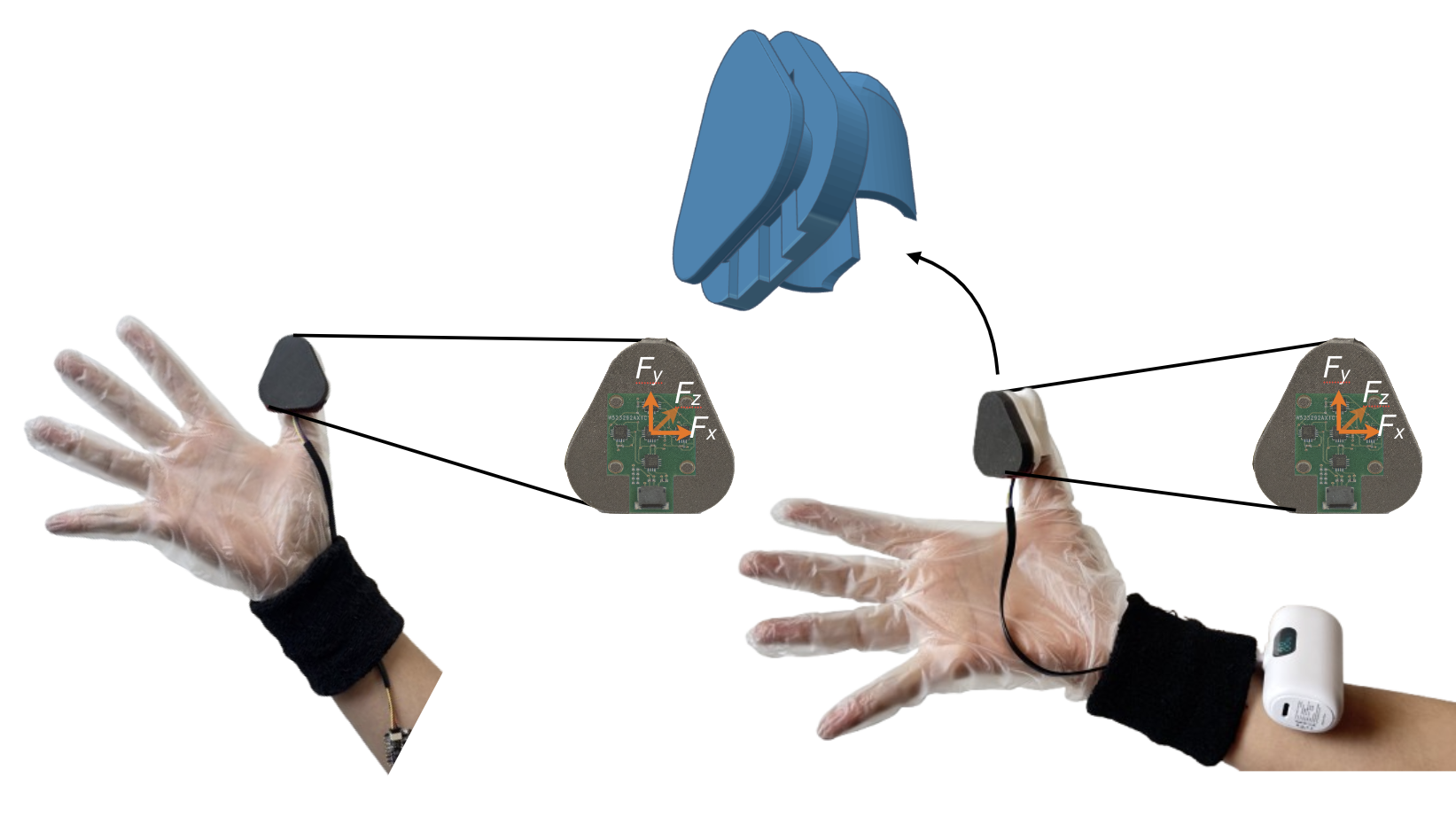}
        \caption{AnySkin augmented glove, worn by a human data-collector. The straightforward electronics of the sensor interface both reduces excessive wiring and also allows for a bluetooth setting (right).}
        \label{fig:human_hardware}
    \end{subfigure}
    \hfill
    \begin{subfigure}[t]{0.45\linewidth}
        \centering
        \includegraphics[width=\linewidth]{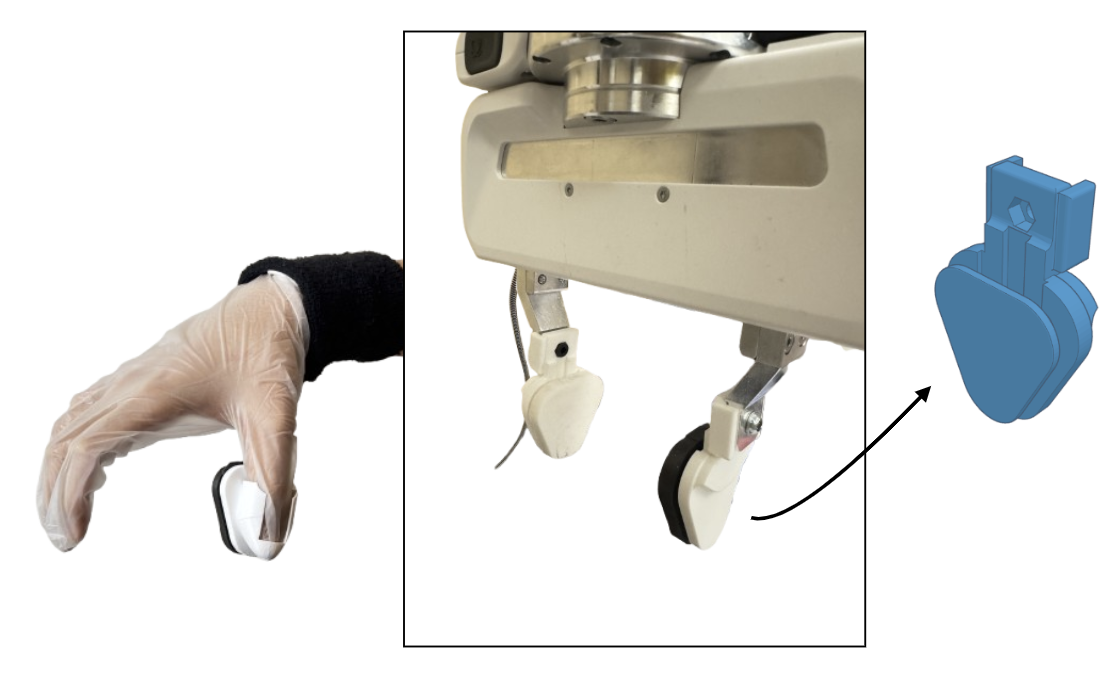}
        \caption{Middle: Franka Panda gripper with AnySkin on one fingertip, emulating the human wearable (left). We attach a plain silicone cap on the other fingertip.}
        \label{fig:robot_hardware}
    \end{subfigure}
    \caption{Hardware setup for human demonstration and robot replication using AnySkin \cite{pattabiraman2024learningprecisecontactrichmanipulation}.}
    \label{fig:hardware_setup}
\end{figure}

\paragraph{Human Hardware Design}
During human data collection, force interactions are captured using a custom ergonomic tactile glove, inspired by AnySkin~\cite{bhirangi2024anyskinplugandplayskinsensing}, which uses 3D-printed magnetometer-based sensors on the underside of the thumb (palm side) to avoid obstructing manipulation. The glove’s transparent design preserves the hand’s natural visual appearance, while an embedded PCB collects force data and streams it via USB to a nearby desktop computer. A schematic of the glove design is shown in Figure~\ref{fig:human_hardware}. While each magnetometer captures full 3D force vectors, we aggregate the force reading by taking the norm of the center magnetometer's force vector. Since the force readings are streamed at 200 fps while the camera readings are 30 fps, we aggregate force readings across time to produce an effective fps of nearly 30. We provide the sensor norm to applied force (Newton) mapping in Figure~\ref{fig:norm-force}.

\paragraph{Robot Hardware Design}
During robot deployment, we 3D print custom gripper tips for the robot end-effector with a mount for fixing the AnySkin tactile sensors. The tactile sensor is only mounted on one of the gripper jaws to emulate the setup on the tactile glove. This ensures a one-to-one correspondence between the sensors used for human data collection and deployment.  The robot gripper force sensing design is shown in Figure~\ref{fig:robot_hardware}. 


\subsection{Embodiment Agnostic Scene Representation}
The human hand motion data from tactile gloves is converted into a point-based representation to enable robot policy learning from human demonstrations.

\subsubsection{Human-to-Robot Embodiment Transfer} For each time step $t$ of a human video, we use Mediapipe~\cite{mediapipe} to extract image key points $p_h^t$ on the human hand. Using point triangulation, the corresponding hand key points from two fixed, calibrated camera views are projected to 3D hand key points $\mathcal{P}_h^t$.  We use point triangulation for 3D projection due to its higher accuracy as compared to sensor depth from the camera~\cite{haldar2025pointpolicyunifyingobservations}. The robot position $\mathcal{R}_{pos}^t$ is computed as the midpoint between the tips of the index finger and thumb in $\mathcal{P}_h^t$. The robot orientation $\mathcal{R}_{ori}^t$ is computed as

\begin{equation}
    \label{eq:orientation}
    \begin{aligned}
        \Delta \mathcal{R}_{ori}^t &= \mathcal{T}(\mathcal{P}_h^{0}, \mathcal{P}_h^{t}) \\
        \mathcal{R}_{ori}^t &= \Delta \mathcal{R}_{ori}^t \cdot \mathcal{R}_{ori}^{0}
    \end{aligned}
\end{equation}

where $\mathcal{T}$ computes the rigid transform between hand key points on the first frame of the video, $\mathcal{P}_h^{0}$, and $\mathcal{P}_h^{t}$. The robot end effector pose is then represented at $T_r^t \leftarrow\{\mathcal{R}_{pos}^t, \mathcal{R}_{ori}^t\}$.
Finally, the robot pose $T_r^t$ is converted to $N$ robot key points through a set of $N$ rigid transformations $T$ about the computed robot pose such that

\begin{equation}
    (\mathcal{P}_r^t)^i = T_r^t \cdot T^i, ~~\forall i \in \{1, ..., N\}
\end{equation}

The robot's gripper state $\mathcal{R}_g$ is considered closed when the distance between the tip of the index finger and thumb is less than 7cm, otherwise open. The continuous force value measured for each step, $\mathcal{R}_f^t$, is also included in the robot state. This process has been illustrated in Figure~\ref{fig:ftf_method}.

\subsubsection{Scene Key Point Representation}
The environment is represented as key points through sparse human annotations, following prior work~\cite{levy2024p3poprescriptivepointpriors, haldar2025pointpolicyunifyingobservations}. Given a single demonstration frame, a human user annotates semantically meaningful key points on task-relevant objects in the scene. Using DIFT~\cite{tang2023emergentcorrespondenceimagediffusion}, an off-the-shelf semantic correspondence model, the annotations are propagated to the first frames of all other demonstrations, minimizing human effort. For each demonstration, Co-Tracker~\cite{karaev2024cotrackerbettertrack}, an off-the-shelf point tracker, then tracks the initialized key point through each trajectory, efficiently handling occlusions and maintaining temporal consistency. To obtain 3D object key points, we triangulate the tracked key points from the two camera views, grounding them in the robot’s base frame. During inference, DIFT is used to localize keypoints in the first frame, after which Co-Tracker tracks them during execution. This approach leverages large pre-trained vision models to generalize across novel object instances and scenes without additional training, requiring only a single frame of user input per task. 

\subsection{Policy Learning}
For policy learning, we use a transformer policy architecture~\cite{haldar2024bakuefficienttransformermultitask, haldar2025pointpolicyunifyingobservations} that takes as input the robot points $\mathcal{P}_r$ and object points $\mathcal{P}_o$ along with the binarized gripper state $\mathcal{R}_g$ and continuous force value $\mathcal{R}_f$. Since the gripper state and force value are 1D and the points are 3D, we repeat the value 3 times when appending to the point tracks to ensure dimensional consistency. A history of observations for each key point is flattened into a single vector and encoded using a multilayer perceptron (MLP) encoder. Each encoded point track and the history of gripper and force values are fed as a separate token into the transformer policy, which predicts the future tracks for each robot point $\mathcal{\hat P}_r$, the robot gripper state $\mathcal{\hat G}_r$, and future gripper force predictions $\mathcal{\hat F}_r$. using a deterministic action head. Following prior works in policy learning~\cite{zhao2023learning, chi2023diffusion}, we use action chunking with exponential temporal averaging to ensure temporal smoothness of the predicted point tracks. The policy is trained using a mean squared error loss. The transformer is non-causal in this scenario, and the training loss is only applied to the robot point tracks. 




\subsection{Inference}

\begin{wrapfigure}{l}{0.48\textwidth}  
\vspace{-3em}
\begin{minipage}[t]{0.48\textwidth}
\begin{algorithm}[H]
\caption{\textsc{ForceFeedbackGripperControl}$(\hat{F}_t)$}
\begin{algorithmic}[1]
\STATE Initialize $\tau \leftarrow 0$
\REPEAT
    \STATE $\Delta g_t^{\tau} = k \cdot (\hat{F}_t - F_t^{\tau})$
    \STATE $g_t^{\tau+1} = g_t^{\tau} + \Delta g_t^{\tau}$
    \STATE Execute gripper action $g_t^{\tau+1}$
    \STATE Read $F_t^{\tau+1}$ from AnySkin
    \STATE $\tau \leftarrow \tau + 1$
\UNTIL{$||\hat{F}_t - F_t^{\tau}|| \leq \epsilon$}
\end{algorithmic}
\end{algorithm}
\end{minipage}
\vspace{-1.5em}
\end{wrapfigure}

\paragraph{Robot pose from predicted key points} The predicted robot points $\mathcal{\hat P}_r$ are mapped back to the robot pose using constraints from rigid-body geometry. We first consider the key point corresponding to the robot's wrist $\mathcal{\hat P}_r^{wrist}$ as the robot position $\mathcal{\hat R}_{pos}$. The robot orientation $\mathcal{\hat R}_{ori}$ is computed using Eq.~\ref{eq:orientation} considering $\mathcal{R}_{ori}^{0}$ is fixed and known. Finally, the robot pose $\mathcal{\hat R}_{pose}$ is defined as $(\mathcal{\hat R}_{pos},~\mathcal{\hat R}_{ori})$.



\begin{minipage}[t]{\textwidth}
\begin{algorithm}[H]
\caption{\method Policy Inference}
\begin{algorithmic}[1]
\STATE Obtain object keypoints on first frame using DiFT on annotated dataset frame.
\FOR{t in rollout}
    \STATE Compute action chunk $(\hat{\tilde{a}}_t,...,\hat{\tilde{a}}_{t+H}) = \pi(a|s_t)$ and obtain $\hat{a}_t$ with temporal aggregation.
    \STATE Parse action: $(\hat{F}_t, \hat{g}_t, \hat{a}^{eef}_t) \leftarrow \hat{a}_t$
    \IF{$\hat{g}_t > closethreshold$}
        \STATE Call \textsc{ForceFeedbackGripperControl}$(\hat{F}_t)$
    \ELSIF{$\hat{g}_t < openthreshold$}
        \STATE Open gripper
    \ENDIF
    \STATE Execute $\hat{a}^{eef}_t$ on robot
    \STATE Read next state $s_{t+1}$ using Co-Tracker
\ENDFOR
\end{algorithmic}
\end{algorithm}
\end{minipage}

\paragraph{Inference-time PD force controller} 
To deploy the tactile policy on the robot arm, we need a means for the robot gripper exerting the force predicted by the policy at each step. For this, we design an outer-loop PD controller that adjusts the target gripper closure setpoints to stabilize the measured forces. If at some timestep $t$, the policy predicts a force $\hat{F}_t$ to be applied, the controller is:
\begin{equation}
    \Delta g_{t}^{\tau} = k \cdot (\hat{F}_t - F_t^{\tau}) 
\end{equation}
where $\tau$ is the inner loop timestep of the PD controller and $F_t^{\tau}$ is the force read by the robot at timestep $t$ of the policy and timestep ${\tau}$ of the controller. At each step the gripper closure is updated as $g_t^{\tau+1} = g_t^{\tau} + \Delta g_{t}^{\tau}$. \par\vspace{1em} The PD controller runs until the convergence condition $||\hat{F}_t - F_t^{\tau}|| < \epsilon$. After the controller converges to the desired $\hat{F}_t$, the policy predicts the next action for step $t+1$. We find $k=0.001$ and $\epsilon=5$ to work well across all tasks. Finally, the action $\mathcal{\hat A}_r = (\mathcal{\hat R}_{pose}, \mathcal{\hat G}_r, g_t)$ is executed on the robot using end-effector position control at a 6Hz frequency.

\section{Experiments}
Our experiments are designed to answer the following questions: (1) How well does \method work for learning force-sensitive tasks compared to baselines that learn to use human force data passiveley? (2) How does \method compare to tactile policy learning methods that learn from robot teleoperation data? (3) Is \method robust to test-time disturbances?

\textbf{Experimental Setup}   
We evaluate \method on a Franka Panda robot, operating in a real-world tabletop manipulation environment. Two Intel RealSense D435 cameras are mounted to provide third-person RGB images to our policy. For baselines we also collect 30 demonstrations on the Franka robot per task using a VR-based teleoperation framework \cite{iyer2024openteachversatileteleoperation}. Demonstrations are recorded at 20Hz and subsampled to approximately 6Hz. For methods outputting robot actions, we use absolute actions with orientation represented with a 6D rotation representation \cite{rotation}.

\begin{figure}[t]
    \centering
    \includegraphics[width=\linewidth]{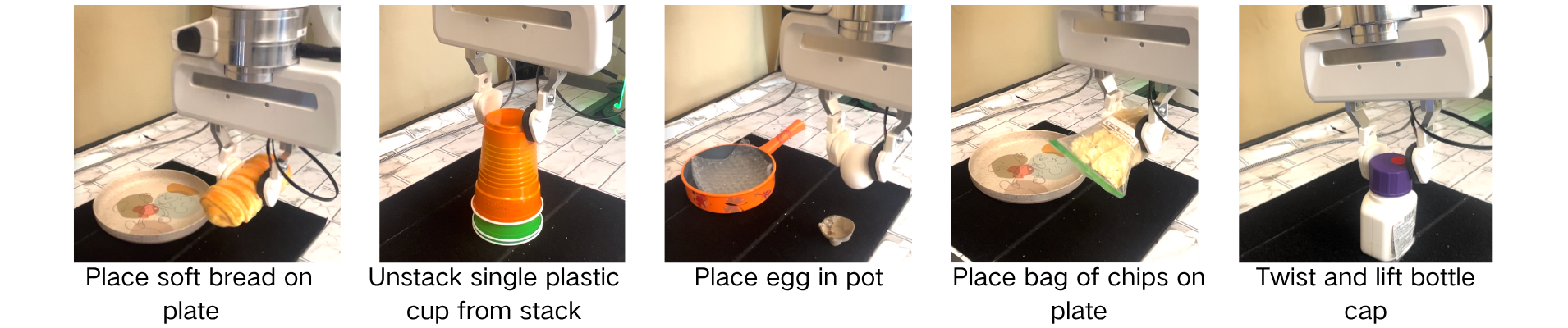} 
    \caption{Manipulation tasks evaluated with \method}
    \label{fig:all_tasks.png}
\end{figure}

\textbf{Task Descriptions} Our manipulation tasks involve variations designed to evaluate the scope of force-sensitive manipulation capabilities achievable with \method. The tasks are depicted in Figure \ref{fig:all_tasks.png}. We provide a description of each task below. For each task, we collect 30 human demonstrations and 30 teleoperated demonstrations.

    a) \texttt{Place soft bread on plate}: The robot arm picks up a highly deformable piece of bread from the table and places it on the plate without crushing it. The positions of the bread are varied for each evaluation.  \\
b) \texttt{Unstack single plastic cup from stack}: The robot arm unstacks a single plastic cup from a stack of 3 upside down cups on the table.  \\
c) \texttt{Place egg in pot}: The robot arm gently picks up and places an egg in a pot without crushing it. The position of the egg is varied for each evaluation. \\
d) \texttt{Place bag of chips on plate}: The robot arm picks and places a transparent bag of chips into a plate without crushing any of the chips inside. \\
e) \texttt{Twist and lift bottle cap}: The robot arm twists and lifts the cap of a bottle to remove it.  \\

\textbf{Baselines}
We compare \method with 5 baselines - \textit{Tactile Point Policy} \cite{haldar2025pointpolicyunifyingobservations}, \textit{Continuous-Gripper Tactile Point Policy}, \textit{FTF + Tactile P3-PO }\cite{levy2024p3poprescriptivepointpriors}, \textit{Tactile P3-PO}, and \textit{Continuous-Gripper Tactile P3-PO}. We describe each method below. \\
a) \textit{Tactile Point Policy} \cite{haldar2025pointpolicyunifyingobservations} performs behavior cloning from point tracks extracted from human data as well as force readings from the tactile glove and predicts future tracks which are converted into robot actions. This baseline provides a comparison to methods such as \cite{pattabiraman2024learningprecisecontactrichmanipulation} that use force input to improve the precision of learned policies but in the context of human data. \\
(b) \textit{Continuous-Gripper Tactile Point Policy} is similar to \textit{Tactile Point Policy} but predicts continuous gripper closure. The gripper closure value is measured as the distance between the index and thumb tracked points from the human data re-normalized to the range of the robot gripper. \\ 
(c) \textit{FTF + Tactile P3-PO} extends \textit{Tactile P3-PO} by predicting both robot actions and future contact forces. The model is trained on teleoperated robot data, using force signals collected during teleoperation as input, and outputs predicted forces alongside actions. This baseline evaluates whether incorporating force prediction improves control performance in robot teleoperation setting and compares the utility of robot-collected versus human-collected force data.\\
(d) \textit{Tactile P3-PO} \cite{levy2024p3poprescriptivepointpriors} predicts teleoperated robot actions from robot tracks obtained by unprojecting robot and object points of interst into 3D space. The method also inputs force readings from the robot gripper collected during teleoperation into the Transformer policy. This method provides a similar comparison to \textit{Tactile Point Policy} but on teleoperated robot data and ground truth robot actions. \\
(e) \textit{Continuous-Gripper Tactile P3-PO} is similar to \textit{Tactile P3-PO} but predicts continuous gripper closure. The continuous gripper values are obtained directly from the robot gripper during teleoperator using an adaptation to the VR-based teleoperation framework \cite{iyer2024openteachversatileteleoperation} that allows the teleoperator to output continuous gripper closures based on visual feedback during data collection.

\begin{table*}[t]
\centering
\caption{Performance comparison of different gripper action spaces in Human Demo}
\label{table:human_demo}
\renewcommand{\tabcolsep}{3pt}
\renewcommand{\arraystretch}{1}
\begin{tabular}{lcccc}
\toprule
\multicolumn{1}{c}{\textbf{Task}} & \textbf{FTF} & \textbf{Binary Gripper} & \textbf{Continuous Gripper} \\ 
\midrule
\texttt{Place soft bread on plate} & 13/15 & 0/15 & 0/15 \\
\texttt{Unstack cup from stack} & 9/15 & 0/15 & 0/15 (2/15 picked 3 cups) \\
\texttt{Place egg in pot} & 13/15 & 0/15 & 0/15 \\
\texttt{Place bag of chips on plate} & 10/15 & 0/15 & 0/15 \\
\texttt{Twist and lift bottle cap} & 13/15 & 11/15 {\scriptsize (1/15 break gripper pads)} & 0/15 \\
\bottomrule
\end{tabular}
\end{table*}
\begin{table*}[t]
\centering
\caption{Performance comparison of different gripper action spaces in Robot Teleop Demo}
\label{table:robot_teleop}
\renewcommand{\tabcolsep}{4pt}
\renewcommand{\arraystretch}{1}
\begin{tabular}{lccc}
\toprule
\multicolumn{1}{c}{\textbf{Task}} & \textbf{FTF} & \textbf{Binary Gripper} & \textbf{Continuous Gripper} \\ 
\midrule
\texttt{Place soft bread on plate}  & 5/15 & 0/15 & 3/15 \\
\texttt{Unstack cup from stack} & 4/15 & 0/15 {\scriptsize (6/15 picked 3 cups)} & 0/15 {\scriptsize (2/15 picked 2 cups)} \\
\texttt{Place egg in pot} & 0/15 & 0/15 & 0/15 \\
\texttt{Place bag of chips on plate} & 3/15 & 0/15 & 0/15 \\
\texttt{Twist and lift bottle cap} & 9/15 & 12/15 & 8/15 \\
\bottomrule
\end{tabular}
\end{table*}

\textbf{\method is highly effective at learning low-level force control strategies for force-sensitive tasks.} Table \ref{table:human_demo} compares the success rates of \method with baseline methods. \method stands out as the only approach capable of reliably solving tasks that require precise force application within a narrow range. For example, in the \texttt{unstack single plastic cup from stack} task, \method is the only method able to lift the cup without inadvertently lifting others underneath. In contrast, the binary gripper baseline can lift the cup, but it fails to isolate the cup and ends up lifting all three cups, causing task failure. Similarly, for delicate tasks like \texttt{place soft bread on plate}, \texttt{place egg in pot}, and \texttt{place bag of chips on plate}, the binary gripper fails by crushing the object. In the \texttt{twist and lift bottle cap} task, although the binary gripper achieves 11/15 successes by firmly grasping the rigid cap, it often applies excessive force, leading to failures such as lifting the bottle or damaging the gripper pad. In contrast, \method achieves 13/15 successes with more controlled and precise force application.

The robustness of \textit{Tactile Point Policy's} underlying policy allows the binary gripper baseline to occasionally complete the task, albeit without adhering to the force constraints. Continuous grippers, on the other hand, struggle significantly across tasks. This is because relying solely on finger separation and mapping it to the robot is imprecise, leading to excessive oscillations that undermine stable grasps. Fine-grained positional control of the gripper would likely require more data than we collected in this study, due to the complexity of this mapping.

\textbf{\method generally outperforms teleoperation baselines that use force data passively.
} In Table~\ref{table:robot_teleop}, we also provide results comparing \method with baseline gripper strategies trained on robot teleoperation demonstrations. \method generally performs better than the binary and continuous baselines across most tasks, showing that predicting contact force improves performance in the teleoperation setting.
However, force signals collected during teleoperation are less reliable than those from human demonstrations. For instance, in \texttt{place egg in pot} and \texttt{twist and lift bottle cap}, the \method fails to outperform the binary gripper, indicating inconsistencies or noise in the teleoperated force stream.

We also observe that the continuous gripper baseline struggles due to the sample inefficiency of learning precise gripper closure. While it occasionally performs better—for example, picking up two cups instead of all three in the stacking task—it tends to work only for rigid objects like the bottle cap, where the required closure remains consistent. In contrast, it fails on deformable objects like chip bags, where the required gripper behavior varies more between demonstrations and test-time executions.

\begin{wrapfigure}{l}{0.4\textwidth}
\centering
\small 
\caption{Analysis of \method with test time adversarial disturbance}
\label{table:robust}
\vspace{-0.5em} 
\renewcommand{\tabcolsep}{4pt}
\renewcommand{\arraystretch}{1}
\begin{tabular}{p{0.6\linewidth}c}
\toprule
\textbf{Task} & \textbf{\method} \\ 
\midrule
Place bag of chips on plate & 10/15 \\
\bottomrule
\end{tabular}
\end{wrapfigure}


\textbf{\method is robust to test-time disturbances.} In the \textit{placing a bag of chips on a plate} task, we introduce disturbances by physically interacting with the bag by holding it down to the table, pressing on the top during the lift, or slightly reorienting it. As shown in Table \ref{table:robust}, despite the changes in the observed force profiles, \method is able to adapt and still produce the desired forces with a 67\% success rate.

\section{Conclusion and Limitations}
We present \method, a novel framework for learning force-sensitive manipulation from human tactile demonstrations. By leveraging a tactile glove and vision-based hand pose estimation, \method captures rich contact force signals from natural human interactions without relying on teleoperation or robot-collected data. Our system trains a closed-loop policy to predict hand trajectories and desired contact forces, which are then retargeted to a robot using a PD controller that enables precise and robust force control. Through experiments across diverse manipulation tasks, we demonstrate that \method significantly outperforms prior baselines and and remains robust under perturbations. These results highlight the power of modeling human tactile behavior—paving the way for more effective robot learning from human experience. Existing limitations of \method include: 1) Shear forces are aggregated with normal forces leading to loss of directional information. For more dexterous tasks, force components can be collected, learned, and stabilized independently. 2) The data collection infrastructure is limited to a fixed, calibrated camera setting. Leveraging egocentric cameras and using stereo triangulation for 3D point extraction could allow for in-the-wild data collection.


\clearpage
\acknowledgments{This work
was supported by grants from Microsoft, Honda, Hyundai, NSF award 2339096, and ONR award N00014-22-1-2773. LP is supported by the Sloan and Packard Fellowships.}


\bibliography{example}  

\newpage
\section{Appendix}
\begin{figure}[H]
    \centering
    \begin{subfigure}[t]{0.45\linewidth}
        \centering
        \includegraphics[width=\linewidth]{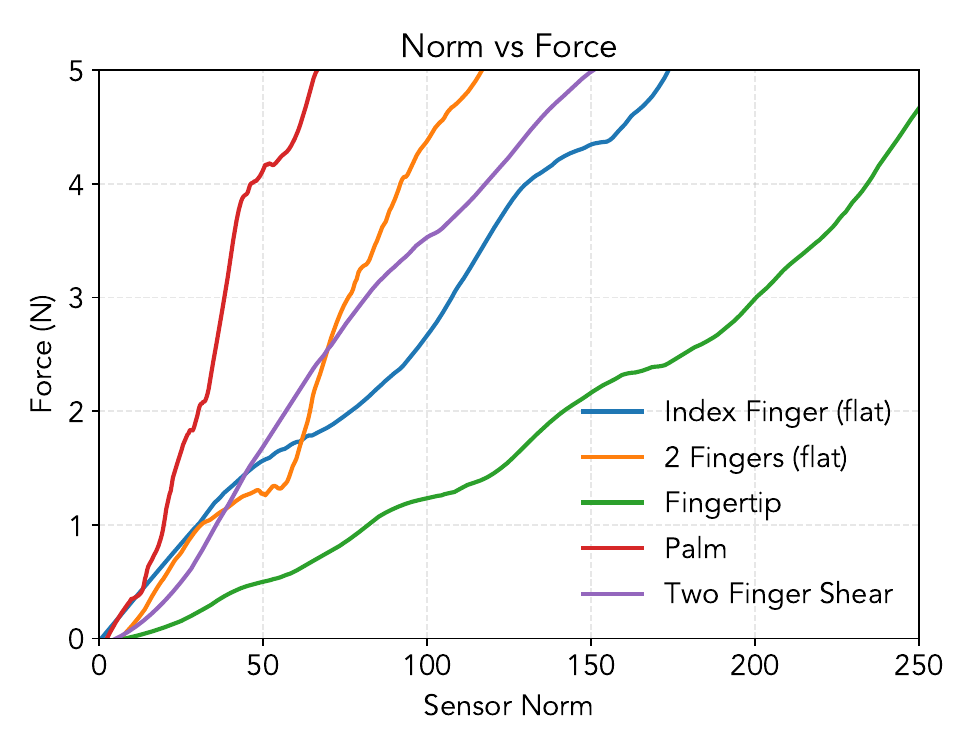} 
        \caption{Transformation between the sensor norm and applied force, for various modes of contact, ranging from low-pressure (Palm) to high pressure (Index Fingertip)}
        \label{fig:norm-plot}
    \end{subfigure}
    \hfill
    \begin{subfigure}[t]{0.45\linewidth}
        \centering
        \includegraphics[width=\linewidth]{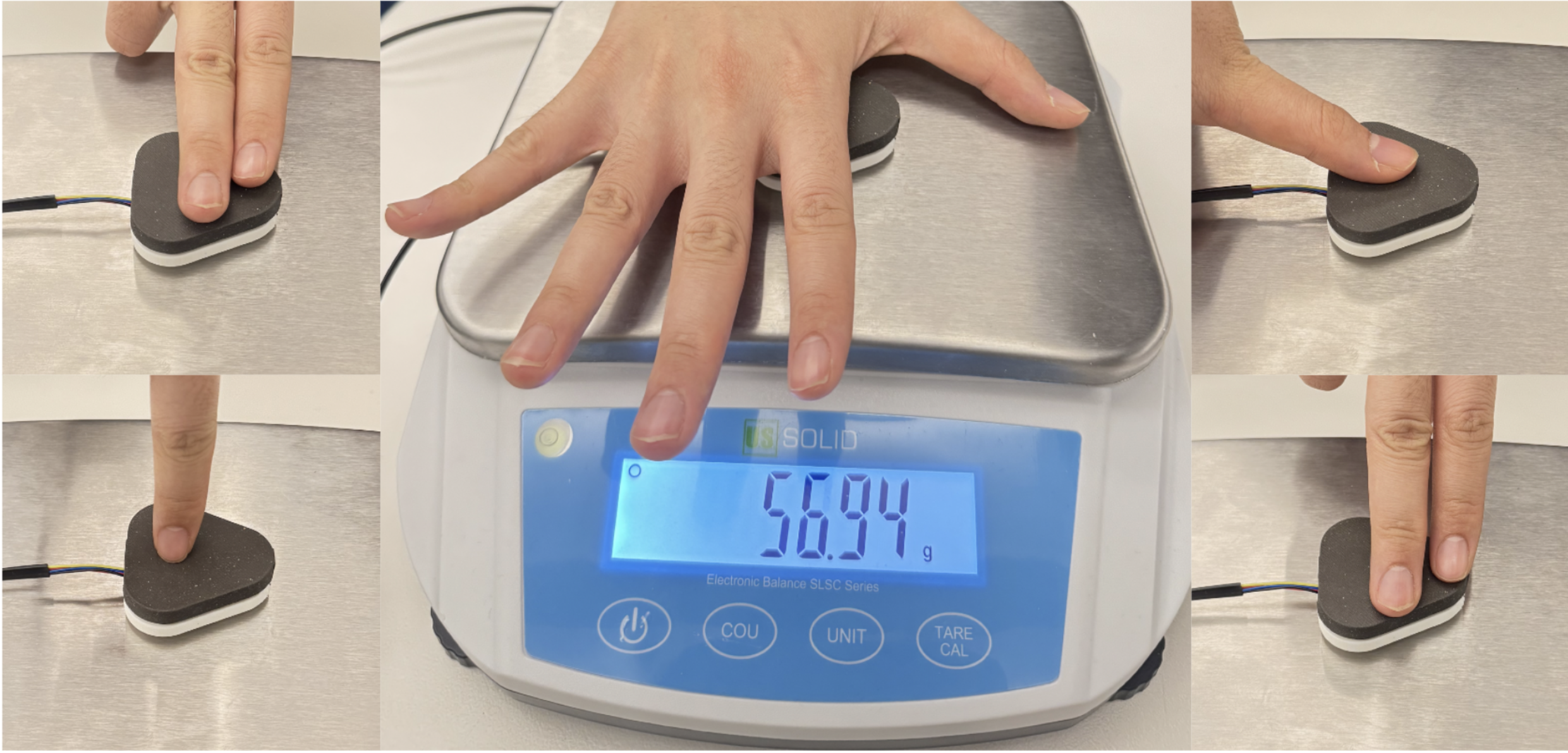} 
        \caption{Different modes of contact for data collection; Two fingers laid flat (top left), index fingertip (bottom left), palm (middle), one finger laid flat (top right) and two fingers pressing at an angle to emulate a combination of both normal and shear forces}
        \label{fig:scale-data}
    \end{subfigure}
    \caption{(Left) Mapping between sensor norm and applied force across various contact modes. (Right) Data collection setup illustrating different types of contact used in the mapping process.}
    \label{fig:norm-force}
\end{figure}

\begin{figure}[H]
    \centering
    \begin{subfigure}[b]{0.75\linewidth}
        \includegraphics[width=\linewidth]{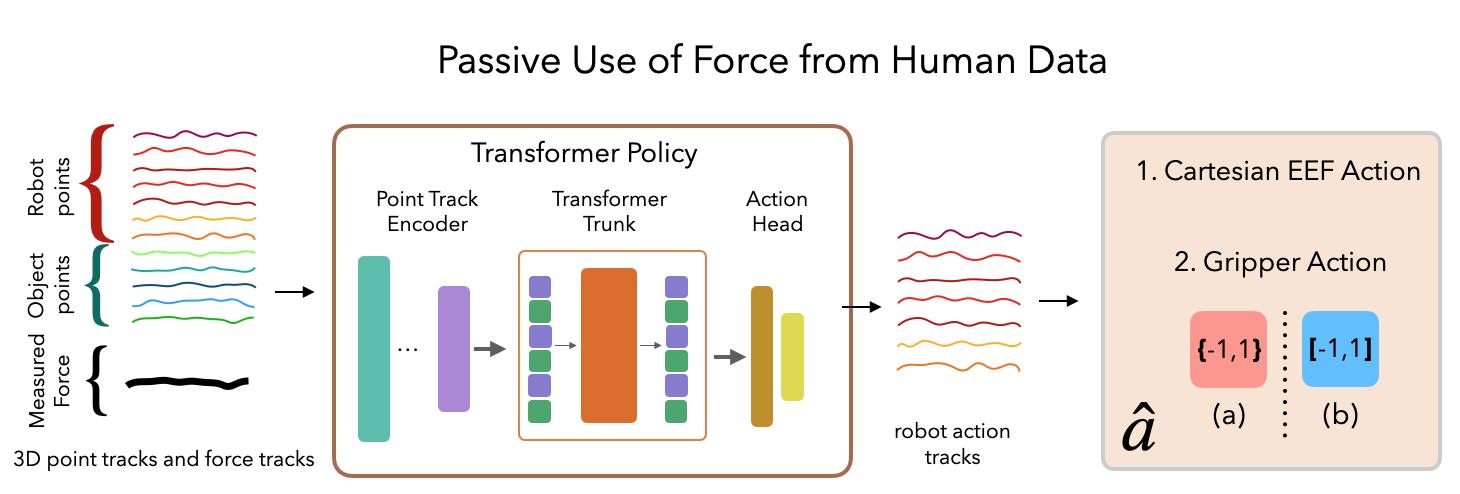}
        \caption{Point track baselines from human data with passive use of force. (a) uses a binary gripper action space by thresholding human hand closure and (b) retargets continuous human hand closure to continuous gripper.}
        \label{fig:sub_human_baselines}
    \end{subfigure}
    \hfill
    \begin{subfigure}[b]{0.75\linewidth}
        \includegraphics[width=\linewidth]{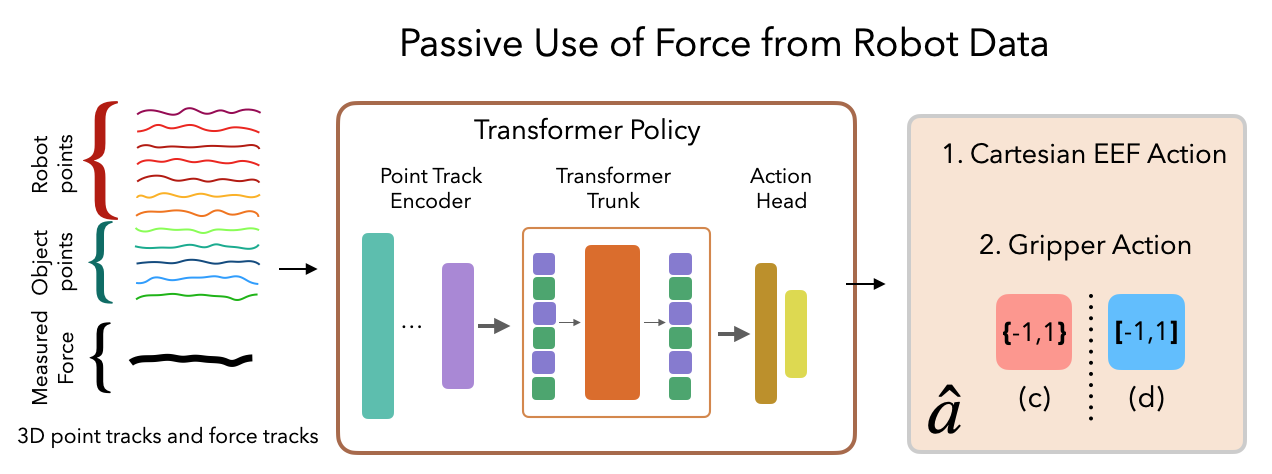}
        \caption{Action imitation baselines from robot data with passive use of force. (c) uses a binary gripper action space and (d) uses continuous gripper action space.}
        \label{fig:sub_robot_baselines}
    \end{subfigure}
    \caption{Comparison of gripper action space across human (left) and robot (right) baselines under passive force conditions.}
    \label{fig:combined_baselines}
\end{figure}

\textbf{Sensor Norm to Force (Newton) conversion} In order to demonstrate a transform between the sensor norm and applied force, we collect data by pressing on an AnySkin sensor mounted on a weighing scale and record synchronous data from both the sensor and the scale streamed through USB at 10Hz. We press the sensor in 5 different manners gradually increasing the force from 0 to 5N, in order to capture different pressures and diverse modes of contact. The sensor norm to applied force comparison is illustrated in Figure~\ref{fig:norm-force}.

\textbf{\method does not require force input to perform effectively.} We implemented a variant of \method that masks the force data fed to the Transformer, constraining the model to predict desired forces solely based on the environment and robot state. We evaluate \method on a simple task \texttt{Lift and hold bread}, involving picking up a soft piece of bread without crushing it and suspending the grasp in the air in Table \ref{table:mask}. This modification did not result in any degradation of force prediction or task performance, suggesting that \method can achieve effective force control even without explicit force input.

\begin{table*}[t]
\centering
\caption{Performance comparison of masked vs umasked force tracks inputted to \method}
\label{table:mask}
\renewcommand{\tabcolsep}{4pt}
\renewcommand{\arraystretch}{1}
\begin{tabular}{lcccc}
\toprule
\multicolumn{1}{c}{\textbf{Task}} & \textbf{Masked Force} & \textbf{Unmasked Force} \\ 
\midrule
\texttt{Lift and hold bread}  & 10/10 & 10/10 \\
\bottomrule
\end{tabular}
\end{table*}

\begin{figure}[H]
    \centering
    \includegraphics[width=\linewidth]{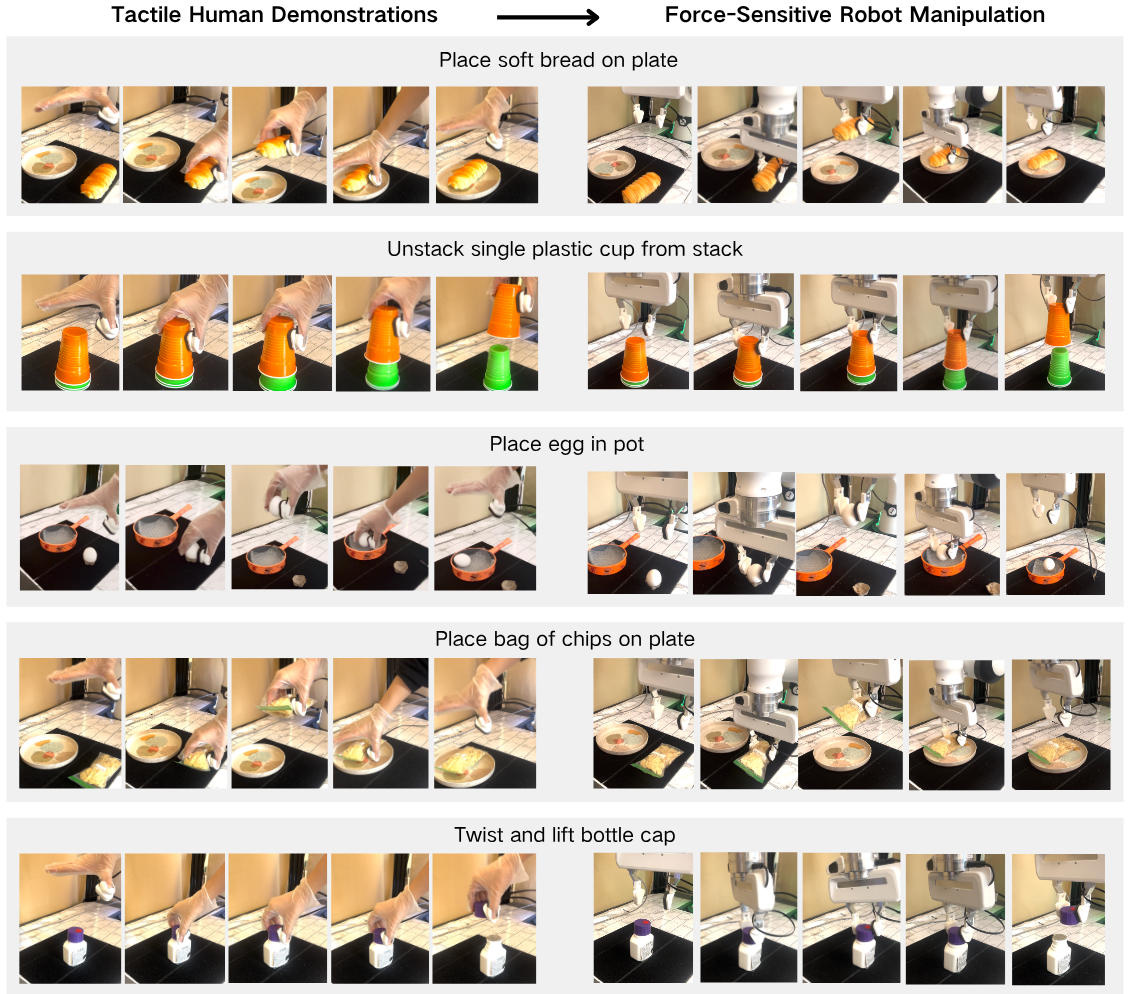}
    \caption{Visual comparison of tactile human demonstrations (left) and force-sensitive robot manipulation rollouts (right) learned from the human demonstrations. From up to down: Place soft bread on plate, unstack single plastic cup from stack, place egg in pot, and place bag of chips on plate.}
    \label{fig:rollouts}
\end{figure}

\end{document}